\documentclass[a4paper,11pt]{article}
\pdfoutput=1
\usepackage{jheppub,bm,bbm}
\usepackage[T1]{fontenc}

\graphicspath{{./Figures/}}
\usepackage{algorithm}
\usepackage[noend]{algpseudocode}

\newcommand{\x}{\mathbf{x}}
\newcommand{\y}{\mathbf{y}}

\newcommand{\ff}{\mathfrak{F}}
\newcommand{\RR}{\mathbb{R}}

\newcommand{\wh}[1]{\widehat{#1}}

\newcommand{\rme}{\mathrm{e}}
\newcommand{\rmd}{\mathrm{d}}
\def\ba#1\ea{\begin{align}#1\end{align}}
\def\mkakko#1{\left(#1\right)}

\def\kkakko#1{\left[#1\right]}
\renewcommand{\r}{\mathbf{r}}
\newcommand{\s}{\mathbf{s}}
\newcommand{\E}{\mathbb{E}}
\newcommand{\bphi}{\bm\phi}
\newcommand{\btheta}{\bm\theta}
\newcommand{\T}{\textsf{T}}
\newcommand{\0}{\mathbf{0}}
\newcommand{\1}{\mathbbm{1}}
\newcommand{\N}{\mathcal{N}}
\newcommand{\V}{\mathbb{V}}
\renewcommand{\c}{\mathbf{c}}
\renewcommand{\k}{\mathbf{k}}

\title{Efficient Bayesian Optimization using Multiscale Graph Correlation}

\author{Takuya Kanazawa}
\affiliation{Research and Development Group, Hitachi, Ltd., Kokubunji, Tokyo 185-8601, Japan}
\emailAdd{takuya.kanazawa.cz@hitachi.com}

\abstract{Bayesian optimization is a powerful tool to optimize a black-box function, the evaluation of which is time-consuming or costly. In this paper, we propose a new approach to Bayesian optimization called GP-MGC, which maximizes multiscale graph correlation with respect to the global maximum to determine the next query point. We present our evaluation of GP-MGC in applications involving both synthetic benchmark functions and real-world datasets and demonstrate that GP-MGC performs as well as or even better than state-of-the-art methods such as max-value entropy search and GP-UCB.}

\begin{document} 
\maketitle
\flushbottom

\section{Introduction}
Global optimization problems involving functions, the evaluation of which is either time-consuming or costly, arise in many fields of science and technology. For instance, in an aeromechanical design problem, the optimal design of a wing is searched for through numerical simulations that are computationally expensive; hence, one may wish to optimize a sequence of simulations to find a better design within a limited computational budget. Bayesian optimization \cite{Kushner1964,Mockus1978,Jones1998} is a well-established methodology to deal with such an optimization problem involving black-box expensive-to-evaluate functions. The two major advantages of Bayesian optimization are sample efficiency and robustness under noisy observations. Bayesian optimization has enabled applications in vast fields such as drug design, automatic machine learning, robotics, and control tasks \cite{Brochu2010tutorial,Shahriari2016,Frazier2018tutorial}. 

A plethora of different schemes for Bayesian optimization have respective pros and cons. We have proposed a new scheme called GP-DC, which utilizes \emph{distance correlation} (DC) \cite{Kanazawa2021}. DC is a generalization of Pearson's correlation coefficient, and it has the ability to detect nonlinear associations between two random variables in a computationally efficient manner \cite{Szekely2007,Szekely2013}. In \cite{Kanazawa2021}, we conducted intensive numerical experiments involving one- and two-dimensional functions to show evidence that GP-DC is on par with or even superior to state-of-the-art methods. 

In this paper, we generalize GP-DC to GP-MGC, which utilizes multiscale graph correlation (MGC) \cite{Vogelstein2019,Shen2020}. MGC, which is a multiscale generalization of DC \cite{Shen2020}, can detect subtle nonlinear relationships between variables of general dimensions in a sample efficient way, and it has enabled applications in brain imaging and cancer genetics \cite{Vogelstein2019}. In this study, we performed a series of benchmark experiments on GP-MGC to show evidence of its superior performance over standard Bayesian optimization schemes. We also tested GP-DC on functions in more than two dimensions, and the results complement those of the previous study \cite{Kanazawa2021} that only focused on one and two dimensions. To solve internal optimization problems of GP-DC and GP-MGC, we used the covariance matrix adaptation evolution strategy (CMA-ES) \cite{Hansen1996,hansen2016cma}, which is a powerful optimization scheme effective for multimodal and/or high-dimensional functions. 

This paper is organized as follows. In section~\ref{sc:pre}, the background of Bayesian optimization with Gaussian processes is provided. In section~\ref{sc:mgc}, the new algorithm utilizing MGC is presented. In section~\ref{sc:exp}, the results of numerical experiments using both synthetic functions and real-world datasets are shown. We conclude in section~\ref{sc:conc}.

\section{Preliminaries}
\label{sc:pre}
\subsection{Gaussian processes}
In a regression task, one is given a finite set of possibly noisy observations $\{(\x_i, y_i)\}_{i=1}^{n}$ of an unknown function $f(\x)$, and one aims to predict a function value at a new point $\x\not\in\{\x_i\}$. In many cases, one wishes to not only predict the function value itself but also to obtain the confidence interval of the prediction. Thus, utilizing a Bayesian framework, which allows a probabilistic inference, is desirable. In Bayesian inference, one needs to supply a prior distribution for the function that represents one's knowledge. In other words, the function space in which one works needs to be specified. To be explicit, let us consider a toy problem about a function $f$ defined over $\RR$. To truncate the infinite-dimensional function space, let us take $1,x$ and $x^2$ as our basis functions. Then, a three-dimensional space of functions may be defined as $g(x)=c_0+c_1x+c_2x^2\equiv\c\cdot\bphi(x)$, where $\bphi(x)\equiv(1,x,x^2)$ and $\c\in\RR^3$. The next step is crucial: assume that $\c$ is sampled from a Gaussian distribution, i.e., $\c\sim\N(\0,\1_3)$. In principle, this assumption allows us to compute any statistical quantity about $g$, e.g., a simultaneous distribution of $g$ at arbitrary $x_1$ and $x_2$: $(g(x_1), g(x_2))\equiv(g_1,g_2)\in\RR^2$. With an elementary calculation, one can show that it is a Gaussian,
\ba
	P(g_1,g_2) & \propto \int \rmd \c\; \rme^{-\c^2/2}
	\delta\mkakko{g_1-\c\cdot \bphi(x_1)}
	\delta\mkakko{g_2-\c\cdot \bphi(x_2)}
	\\
	& \propto \exp\kkakko{
		- \frac{1}{2}(g_1,g_2) K^{-1} \begin{pmatrix}g_1\\g_2\end{pmatrix}
	}\,,
\ea
with the covariance matrix $\displaystyle K\equiv 
\begin{pmatrix}\bphi(x_1)^2 & \bphi(x_1)\cdot \bphi(x_2)
\\\bphi(x_1)\cdot \bphi(x_2) & \bphi(x_2)^2\end{pmatrix}$.
By introducing a \emph{kernel function} $k(x,x')\equiv\bphi(x)\cdot \bphi(x')=1+xx'+x^2x'^2$, we find that $K_{ij}=k(x_i,x_j)$. Therefore, all you need is $k$---the explicit forms of the basis functions are not necessary. Note also that the number of basis functions need not be three but can be made arbitrarily high. It is called \emph{kernel trick} to define a kernel function $k$ with no reference to bases, i.e., the space of functions that may have an infinite number of basis can be accessed through a single function $k$. It is similar to regularization in regression \cite{book_Bishop}. Generally, a naive fit of a model for data, e.g.,~via minimization of least squares $\sum_{i}(y_i-f(\x_i))^2$ is prone to overfitting, and one adds a term that constrains $f$ such as $\int \rmd \x\;|\nabla^2 f(\x)|^2$, which enforces the smoothness of $f$ and mitigates overfitting. The kernel in fact plays a role similar to such regularization terms.

We now state the general definition of the Gaussian process. A stochastic process $\{f(\x)\}_\x$ is called a Gaussian process with a kernel $k(\cdot,\cdot)$ if, for any finite set of points $\{\x_t\}_{t=1:T}$, the simultaneous distribution $(f(\x_1),f(\x_2),\cdots,f(\x_T))\in\RR^T$ is a multivariate Gaussian with $\mathrm{cov}\mkakko{f(\x),f(\x')}=k(\x,\x')$. The choice of the kernel reflects one's prior knowledge on functions. One of the standard choices is the squared exponential (SE) kernel $k(\x,\x')=\exp(-|\x-\x'|^2/\ell^2)$, where $\ell$ stands for the length scale. The SE kernel represents quite smooth functions that are differentiable infinitely many times. 

In a Bayesian framework, one's belief over functions must be modified after making observations $\{(\x_i, y_i)\}_{i=1}^{n}$. The key formulas for Gaussian process regression are \cite{RW_GP_book}
\ba
	\E[f(\x_*)] & = \k_*^\T (K+\sigma^2\1_n)^{-1} \y
	\\
	\V[f(\x_*)] & = k(\x_*,\x_*) - \k_{*}^\T (K+\sigma^2\1_n)^{-1}\k_*
	\\
	\k_* & = \mkakko{k(\x_*,\x_1), \cdots, k(\x_*,\x_n)}^\T
	\\
	K_{ij} & = k(\x_i,\x_j)
	\\
	\y & = (y_1,\cdots,y_n)^\T,
\ea
which specifies the mean and variance of the predictive distribution at a new point $\x_*$. Here, $\sigma^2$ is the noise variance. We note that probabilistic prediction is also possible with tree-based models such as random forests and gradient boosting decision trees, but they are poor extrapolators and tend to be over-confident in regions away from samples. Therefore, we utilize Gaussian processes as our workhorse for probabilistic inference in this paper.

\subsection{Bayesian optimization}
Given a black-box function $f(\x)$ with no information on its functional form or its derivatives, how can we determine $\mathrm{arg\,max}_\x f(\x)$ most efficiently? Bayesian optimization \cite{Kushner1964,Mockus1978,Jones1998} is a powerful technique well suited to solving such a global optimization problem where a function evaluation is either time-consuming or costly and where one wishes to infer the location of the maximum with least observations. An example is automatic machine learning, where one needs to find optimal hyperparameters of a machine learning model such as neural networks within as small a computational budget as possible \cite{Snoek2012}. To solve such a task in Bayesian optimization, we make sequential decisions on where to observe next based on a surrogate model for the function that is built using knowledge of previous observations. This leads to a dilemma between exploration and exploitation: on the one hand, one wants to observe points with large prediction variance to make the surrogate model more accurate (\emph{exploration}), but---on the other---one needs to observe points that are likely to be close to the maximum of the function (\emph{exploitation}). This is resolved by maximizing an \emph{acquisition function} $\alpha(\x)$ that balances exploration with exploitation. Thus, Bayesian optimization consists of running the following loop.
\begin{enumerate}
	\setlength\itemsep{-1mm}
	\item Observe $y_*=f(\x_*)+\varepsilon$ at a new point $\x_*$, where $\varepsilon$ is white noise.
	\item Update the surrogate model by incorporating $(\x_*, y_*)$.
	\item Compute the acquisition function $\alpha(\x)$ using the surrogate model.
	\item Determine $\x_*=\mathrm{arg\,max}_\x \alpha(\x)$.
	\item Return to 1.
\end{enumerate}
Examples of classical acquisition functions are the expected improvement \cite{Mockus1978,Jones1998} and the probability of improvement \cite{Kushner1964}. Newer ones include GP-UCB \cite{Srinivas2012} and knowledge gradient \cite{Frazier2008}. Furthermore, attempts have been made to develop information-theoretic acquisition functions, such as entropy search \cite{Villemonteix2009,Hennig2012}, predictive entropy search \cite{HernandezLobato2014}, output-space entropy search \cite{Hoffman2015}, and max-value entropy search \cite{Wang2017}. We have proposed a new acquisition function GP-DC based on distance correlation \cite{Kanazawa2021}. 

\subsection{Sampling from the posterior distribution}
\label{sc:sam}
After serial observations, sampling of functions from the posterior distribution can be done efficiently using the spectral density of kernel functions and random feature maps \cite{SSGP2010,HernandezLobato2014}. In this study, following the advice of \cite{Snoek2012}, we used the Mat\'{e}rn-5/2 kernel defined as
\ba
	k_{5/2}(\x,\y) = k_{5/2}(r) & \equiv \mkakko{1+\frac{\sqrt{5}r}{\ell}+\frac{5r^2}{3\ell^2}}
	\exp\mkakko{-\frac{\sqrt{5}r}{\ell}}, 
	\quad r\equiv|\x-\y|\,.
\ea
Its spectral density in $D$ dimensions is given by \cite{RW_GP_book}
\ba
	S_\ell(s) & = \frac{\Gamma\mkakko{\frac{D+5}{2}}5^{5/2}\pi^{-\frac{D+11}{2}}}{24 \ell^5} 
	\mkakko{s^2 + \frac{5}{4\pi^2\ell^2}}^{-\frac{D+5}{2}},
\ea
which is normalized as $\displaystyle \int \rmd^D \s ~S_\ell(s)=1$. It holds that $\displaystyle k_{5/2}(r)=\int \rmd^D \s ~\rme^{2\pi i \s\cdot \r}S_\ell(s)$\,. It is straightforward to show that, by sampling $B$ sets of $\s$ and $b$ from the measure $\displaystyle \int \rmd^D\s \; S_\ell(s)\int_0^1\rmd b$ with $B\gg 1$, we have the approximation formula
\ba
	k_{5/2}(\x, \y) & \approx \frac{2}{B}\sum_{i=1}^{B}\cos (2\pi (\s_i\cdot \x+b_i)) \cos (2\pi (\s_i\cdot \y+b_i))
	\\
	& = \bphi^\T(\x) \cdot \bphi(\y),
\ea
where $\bphi$ is a column vector of length $B$ with components $\displaystyle \bphi_i(\x) \equiv \sqrt{\frac{2}{B}}\cos[2\pi (\s_i\cdot \x+b_i)]$. 

In general, the actual kernel is given by $k(\x,\y)=C k_{5/2}(\x,\y)$ with a constant $C>0$, which is a hyperparameter of the model. Then, a sample from the prior distribution can be approximated as $f(\x)\approx \btheta^\T \bphi(\x)$ with $\btheta\sim \N(\0, C \1_B)$. After conditioning on observations $\{(\x_i, y_i)\}_{i=1}^n$, the posterior distribution for $\btheta$ becomes
\ba
	\btheta & \sim \N\bigg(
	C\Phi \frac{1}{C\Phi^\T \Phi+\sigma^2\1_n}\y, 
	C\1_B - C^2\Phi 
	\frac{1}{C\Phi^\T \Phi+\sigma^2\1_n}\Phi^\T
	\bigg),
\ea
where $\Phi \equiv [\bphi(\x_1),\cdots,\bphi(\x_n)]$ is a $B\times n$ matrix.

\section{Bayesian optimization based on MGC}
\label{sc:mgc}
Given a set of previous observations $\{(\x_i,y_i)\}$, how can we determine the point $\x$ that deserves the next observation most? The idea of \cite{Hoffman2015,Wang2017} was to focus on points that have strong correlation with the maximum value $\max_\x f(\x)$. Of course, $\max_\x f(\x)$ itself is not known to us, so we sample many possibilities $\wh{f}_m$ from the posterior distribution, maximize each $\wh{f}_m$, and obtain a probable set of max values $\{\max_{\x}\wh{f}_m(\x)\}_m$. In \cite{Kanazawa2021}, we proposed using DC to quantify a possibly nonlinear correlation between $\{\max_{\x}\wh{f}_m(\x)\}_m$ and $\{f_m(\x)\}_m$. In this paper, we suggest replacing DC with MGC, which is a multiscale generalization of DC and which has better performance for a limited number of samples \cite{Vogelstein2019,Shen2020}. The resulting new algorithm, called GP-MGC, is presented in Algorithm~\ref{alg1}.

\begin{algorithm}[h]
\caption{GP-MGC \label{alg1}}
\begin{algorithmic}[1]
\Require $D_0:=\{(\x_i,y_i)\}_{i\in I}$:~Initial observation data
\For{$t=1,2,\dots$}
\State Update hyperparameters of the Gaussian process using $D_{t-1}$
\State Sample $M$ functions $\{\wh{f}_m\}_{m=1:M}$ from the posterior as described in section~\ref{sc:sam}
\State Find $\wh{\ff}_m:=\max_\x \wh{f}_m(\x)$ for each $m=1,\cdots,M$ via CMA-ES
\State Define $\alpha(\x):=\text{MGC}\mkakko{\{\wh{\ff}_m\}_{m=1:M}, \{\wh{f}_m(\x)\}_{m=1:M}}$
\State Find $\x_t:=\mathrm{arg\,max}_\x \alpha(\x)$ via CMA-ES
\State Observe $f$ at $\x_t$ and obtain $y_t$
\State $D_t\gets D_{t-1}\cup\{(\x_t, y_t)\}$
\EndFor 
\end{algorithmic}
\end{algorithm}

GP-MGC has two steps in which a global optimization has to be performed. The first is line 4, where each sample function is maximized, and the second is line 6, for which the maximum of the acquisition function is searched. The second maximization is encountered in every Bayesian optimization approach, whilst the first maximization is somewhat unique to GP-MGC. We propose solving both optimizations via CMA-ES, a highly efficient algorithm for non-convex global optimization \cite{Hansen1996,hansen2016cma}. Note that the computation in line 4 can be easily parallelized. In this study, we used Scipy \cite{scipy} to compute MGC and Optuna \cite{Optuna} to implement CMA-ES. 

\section{Numerical experiments}
\label{sc:exp}
\subsection{Test on benchmark functions}
\label{sc:bef}
We compared the performance of GP-MGC with baselines: random policy, expected improvement (EI), GP-UCB, max-value entropy search (MES), and GP-DC.

The following standard benchmark functions were used:
\begin{itemize}
	\setlength\itemsep{-1mm}
	\item Michalewicz function (2 dimensions)
	\item Six-hump camel function (2 dimensions)
	\item Hartmann-3 function (3 dimensions)
	\item Ackley function (3 dimensions)
	\item Levy function (4 dimensions)
	\item Hartmann-6 function (6 dimensions)
\end{itemize}
The reader is referred to \cite{SB_website} for more details on these functions. Three random initial points were supplied at the beginning, and afterwards 40 sequential function evaluations were performed in accordance with each policy. The results of 30 random initial conditions were averaged to enable a stable comparison among methods. We measured the performance by \emph{regret} at step $t$ defined by $\displaystyle \max_\x f(\x) - \max \{y_i\}_{i=1}^{t}$, which represents the gap between the true maximum and the maximum of what has been observed so far. Here, we assumed that observation noise was absent. 

\begin{figure}[p]
	\centering
	\includegraphics[width=.45\textwidth]{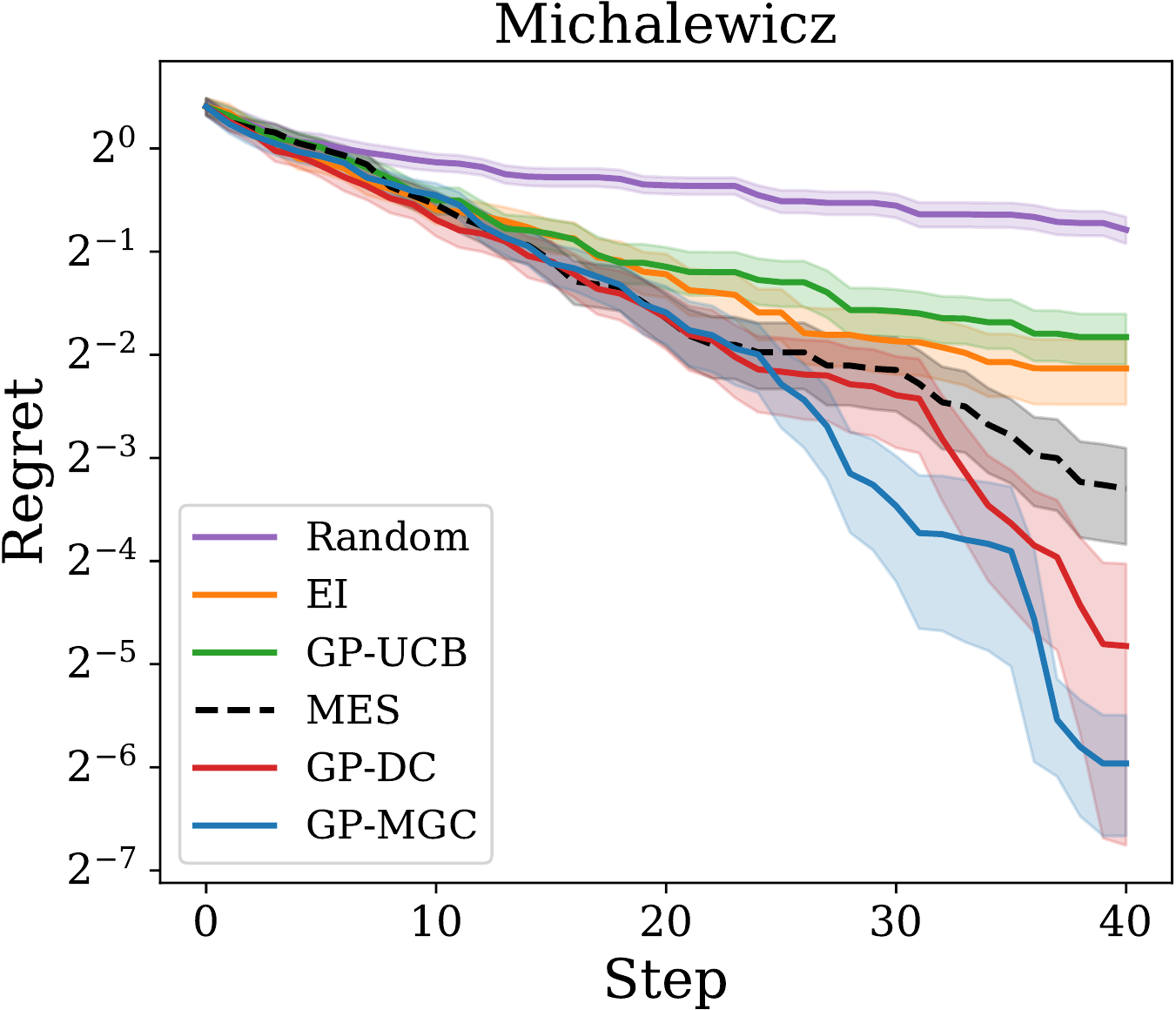}
	\quad 
	\includegraphics[width=.45\textwidth]{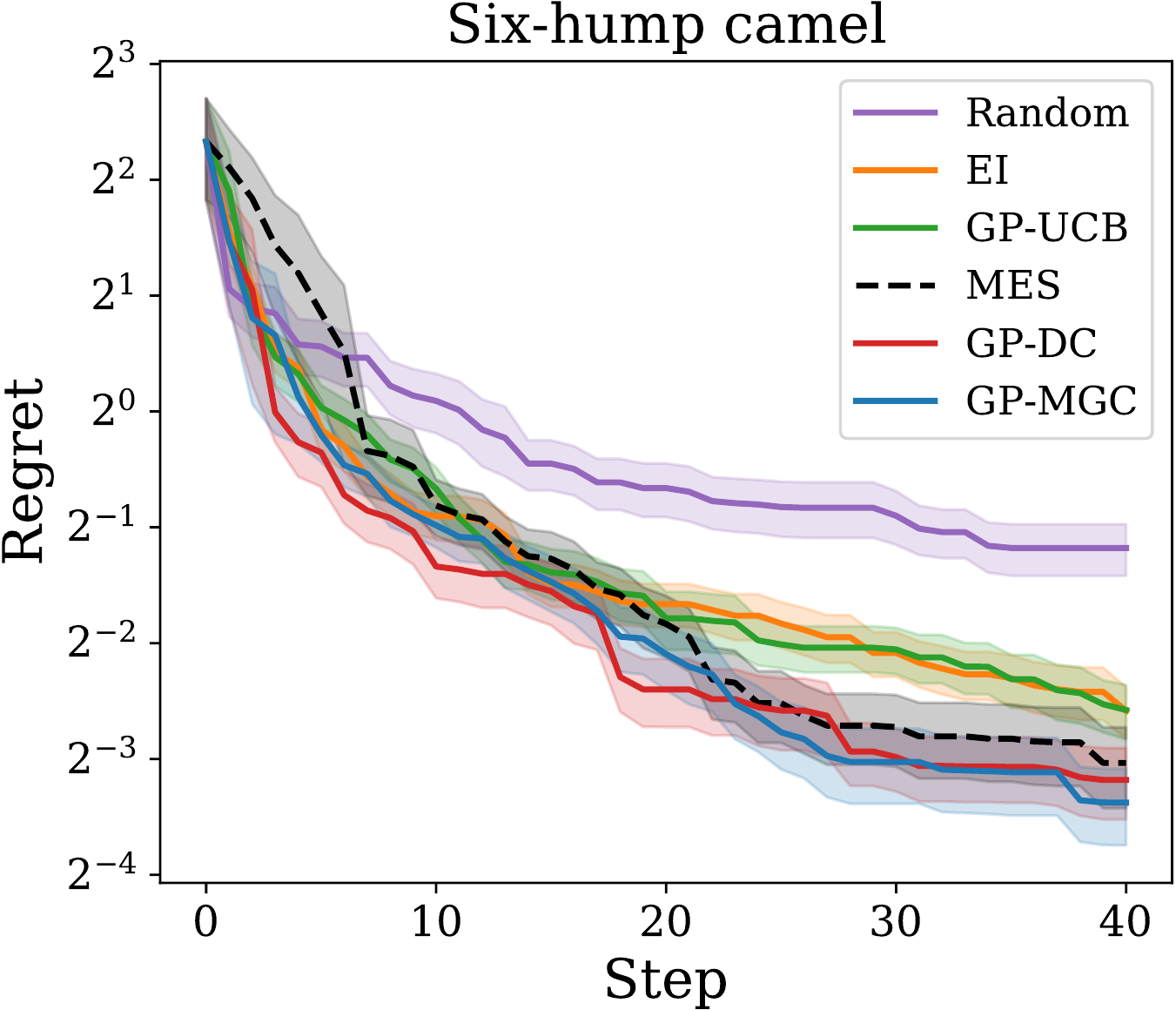}
	\vspace{5mm}\\
	\centering
	\includegraphics[width=.45\textwidth]{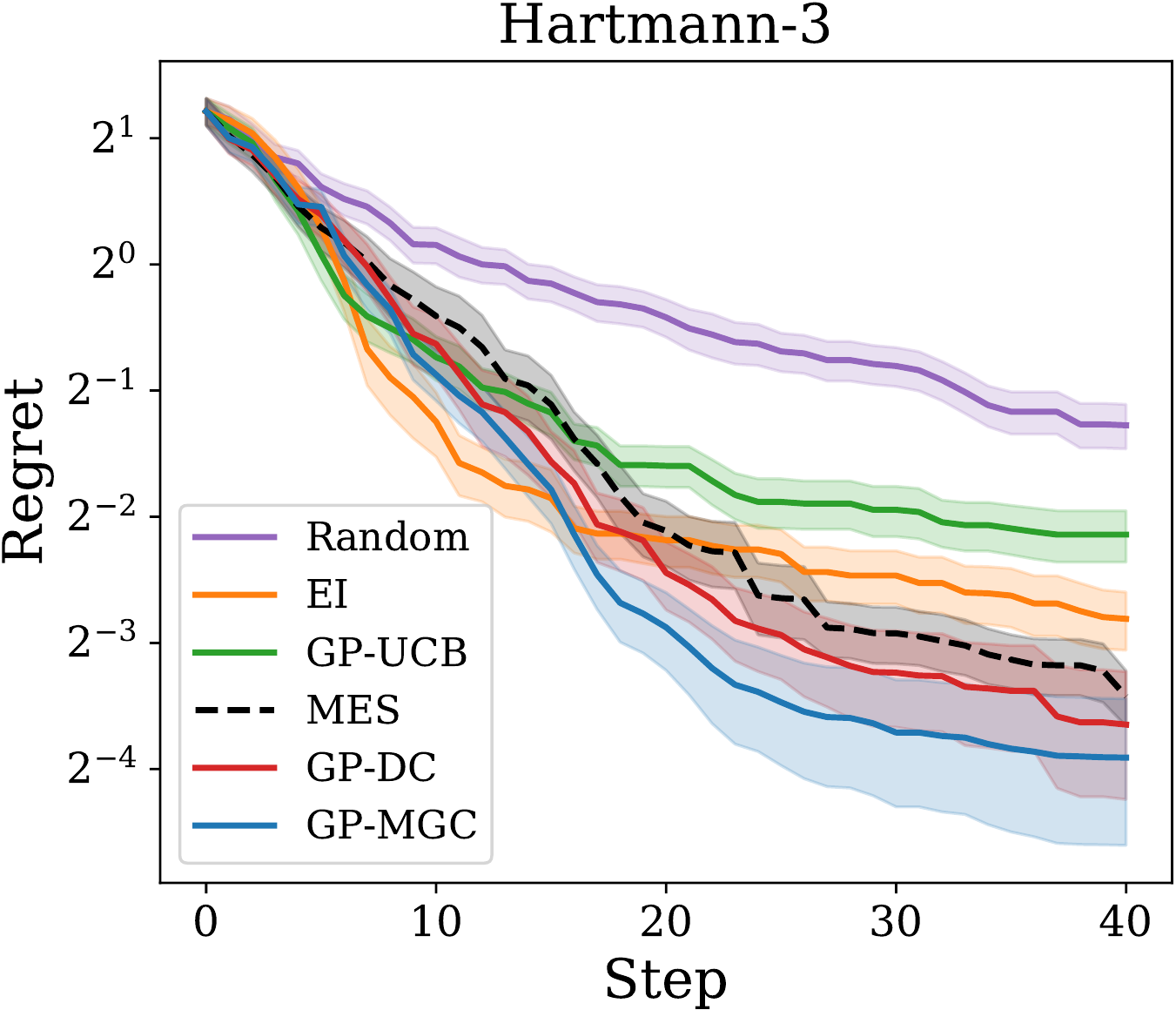}
	\quad 
	\includegraphics[width=.45\textwidth]{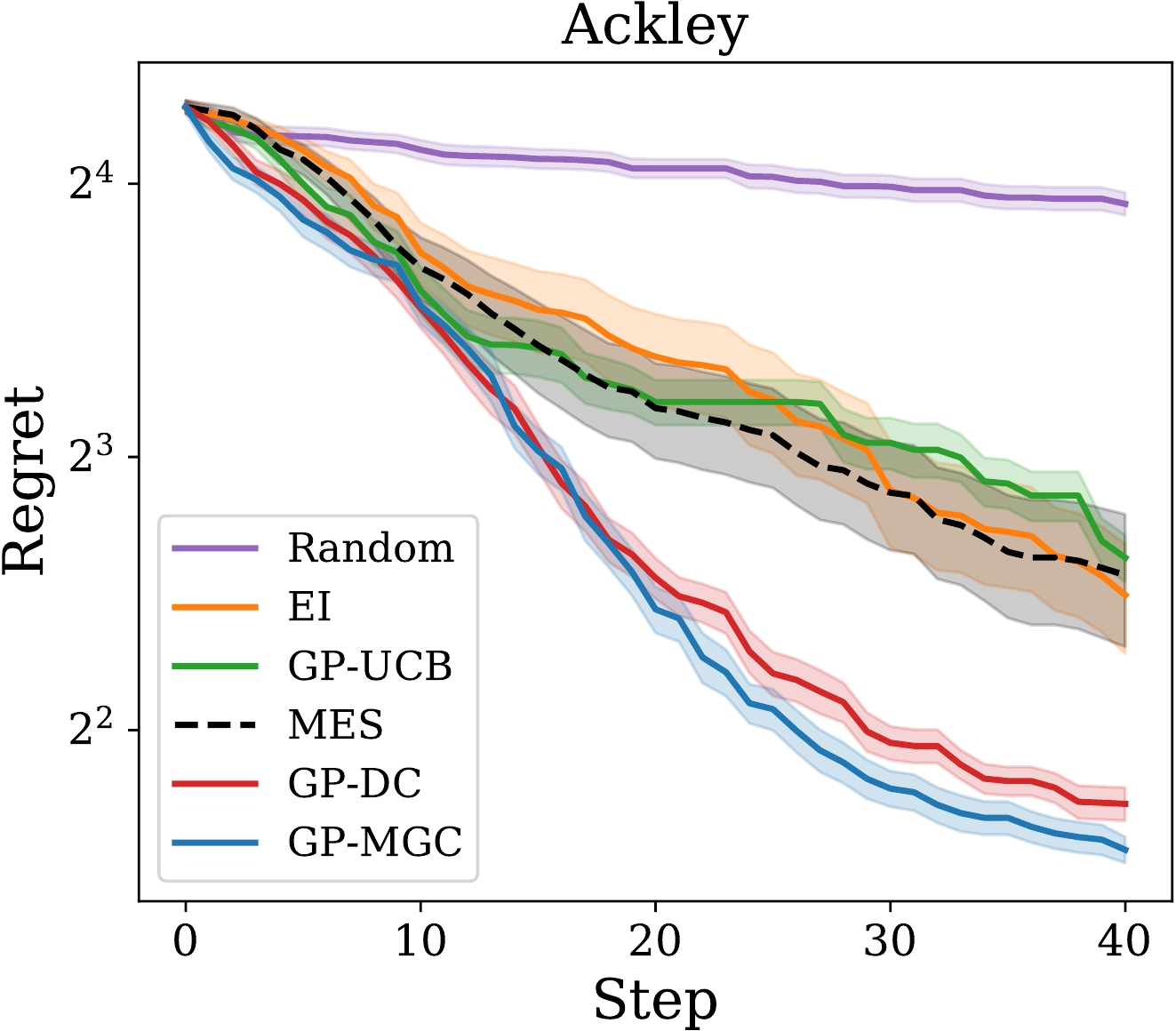}
	\vspace{5mm}\\
	\centering
	\includegraphics[width=.45\textwidth]{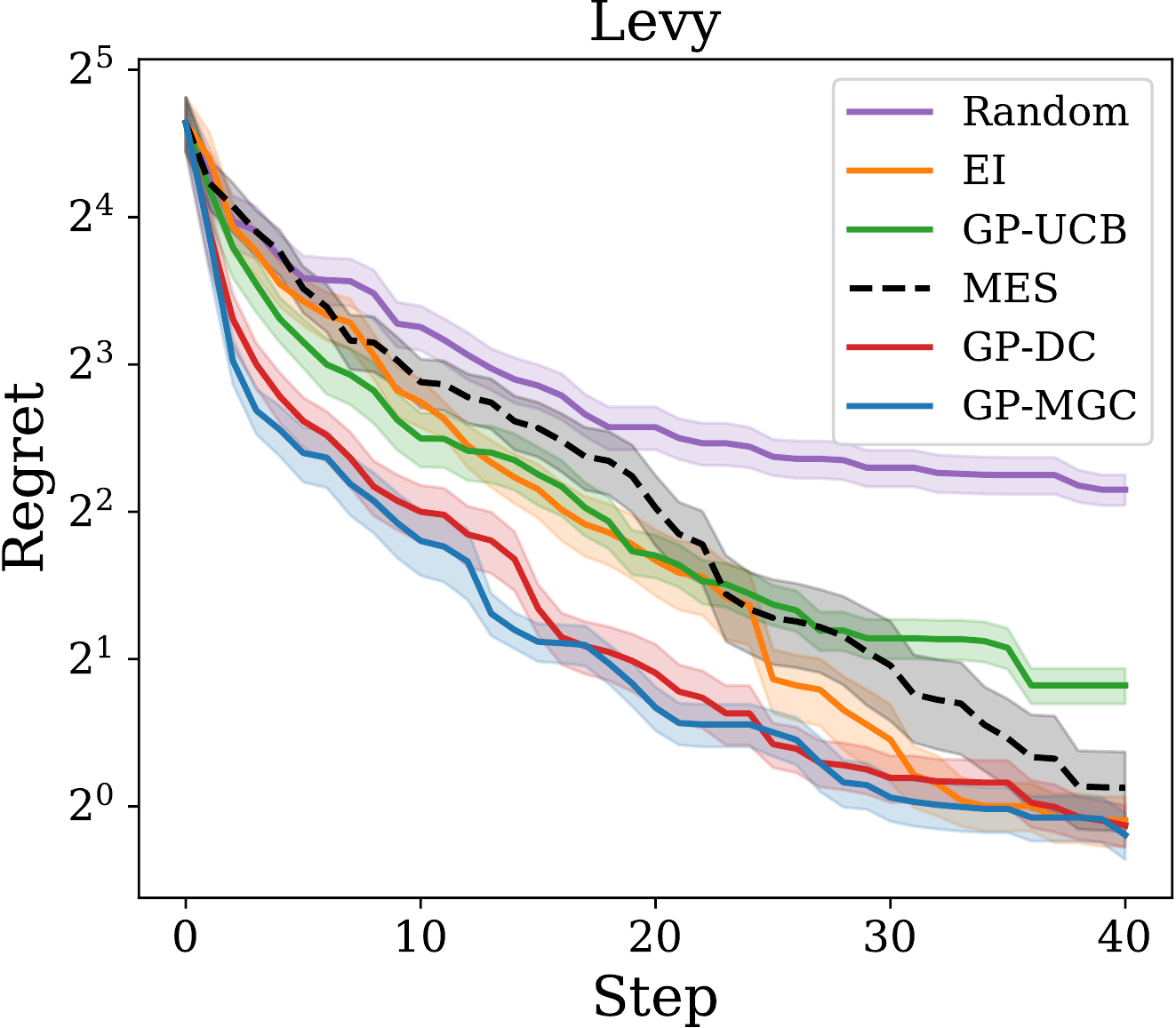}
	\quad 
	\includegraphics[width=.45\textwidth]{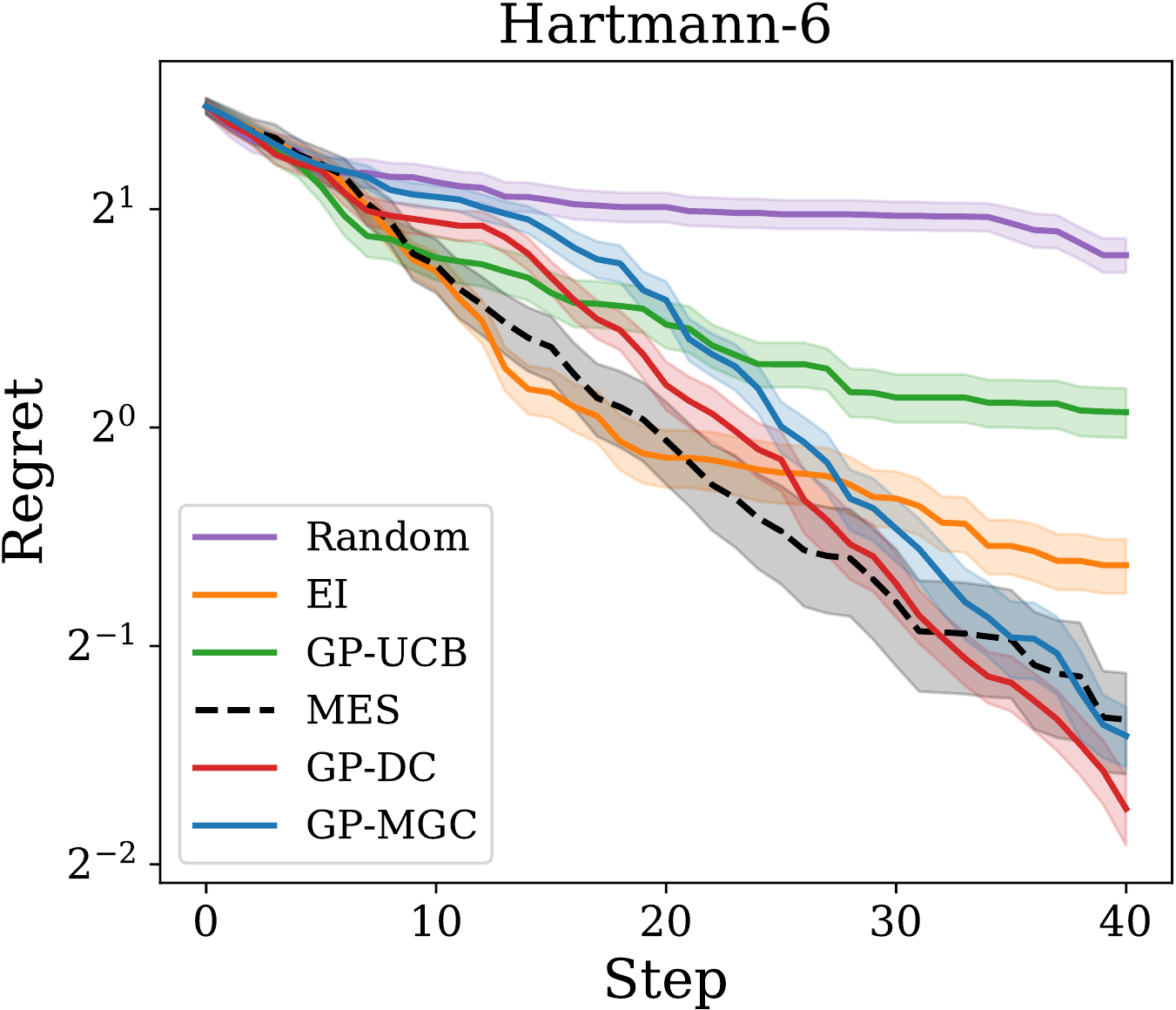}
	\caption{\label{fg:bench}Benchmark results for six test functions. The average regret (on log scale) is plotted as a function of steps. The color band represents one standard deviation of the mean.}
\end{figure}

The numerical results are summarized in figure~\ref{fg:bench}. GP-MGC tended to perform slightly better than GP-DC in general. GP-DC and GP-MGC outperformed the other methods, except for Hartmann-6 for which MES seemed to perform best up until step 30, after which GP-DC won over MES. In table~\ref{tb:bench}, cumulative regret is shown for each method and each test function to help a quantitative comparison. 

Overall, we found that GP-MGC is at least as good as state-of-the-art Bayesian optimization methods. Note that all these benchmark functions are highly multimodal and exceptionally difficult to optimize, implying that GP-MGC could become a powerful tool to solve practical optimization problems in our daily life.
\begin{table}[h]
	\hspace{-3mm}
	\scalebox{0.85}{
	\begin{tabular}{|c|c|c|c|c|c|c|c|c|c|c|c|c|}
		\hline 
		& \multicolumn{2}{c|}{Michalewicz}
		& \multicolumn{2}{c|}{Six-hump camel}
		& \multicolumn{2}{c|}{Hartmann-3}
		& \multicolumn{2}{c|}{Ackley}
		& \multicolumn{2}{c|}{Levy}
		& \multicolumn{2}{c|}{Hartmann-6}
		\\\hline 
		Steps &1--20&21--40&1--20&21--40&1--20&21--40&
		1--20&21--40&1--20&21--40&1--20&21--40
		\\\hline 
		Random &18.7&13.5&22.3&10.3&24.4&10.9&350&318&
		201&99.5&43.9&38.4
		\\
		EI &14.5&5.62&15.8&4.73&{\bf 14.4}&3.55&284&155&
		158&31.0&{\bf 33.0}&15.5
		\\ 
		GP-UCB &15.1&6.93&17.0&4.59&15.8&5.13&260&163&
		142&45.4&36.6&22.9
		\\ 
		MES &14.2&3.93&23.0&3.15&17.0&2.78&270&146&
		177&39.3&34.4$'$&{\bf 12.0}
		\\
		GP-DC &{\bf 13.3}&3.16$'$&{\bf 13.0}&2.77$'$&15.9&2.22$'$&231$'$&82.5$'$&
		96.6$'$&24.4$'$&37.7&12.7$'$
		\\
		GP-MGC &14.0$'$&{\bf 2.42}&14.6$'$&{\bf 2.65}&15.0$'$&{\bf 1.63}&{\bf 227}&{\bf 73.3}&
		{\bf 84.8}&{\bf 22.8}&41.2&15.2
		\\\hline 
	\end{tabular}
	}
	\caption{\label{tb:bench}Cumulative regret (lower is better) for six benchmark functions. The best score is in bold, and the second best is represented by prime $(')$.}
\end{table}

\subsection{Test on real-world datasets}
Next, we turn to the problem of optimal hyperparameter search in machine learning. We utilized support vector regression implemented in scikit-learn \cite{scikit-learn}. The tuned hyperparameter and the search domain are $C\in[10^{-2},10^3]$, $\text{gamma}\in[10^{-3},1]$, and $\text{epsilon}\in[10^{-4},1]$, respectively. We used ten-fold cross validation and evaluated the goodness of hyperparameters using the coefficient of determination $R^2$. As in section~\ref{sc:bef}, at the beginning of each simulation, three random points were given as initial data, followed by 40 sequential function evaluations. We took the average of $R^2$ over 30 independent random initial conditions. 

The following three real-world datasets were used for our benchmarking.
\begin{itemize}
	\setlength\itemsep{-1mm}
	\item Boston housing dataset \cite{Harrison1978,bostonurl}. Number of instances:~$506$, number of features:~$13$.
	\item Concrete compressive strength dataset \cite{cheng1998,concreteurl}. Number of instances:~1030, number of features:~8.
	\item Quantitative structure activity relationships (QSAR) fish toxicity dataset \cite{Cassotti2015,fishurl}. Number of instances:~908, number of features:~6.
\end{itemize}

\begin{figure}[tb]
	\centering
	\includegraphics[width=.45\textwidth]{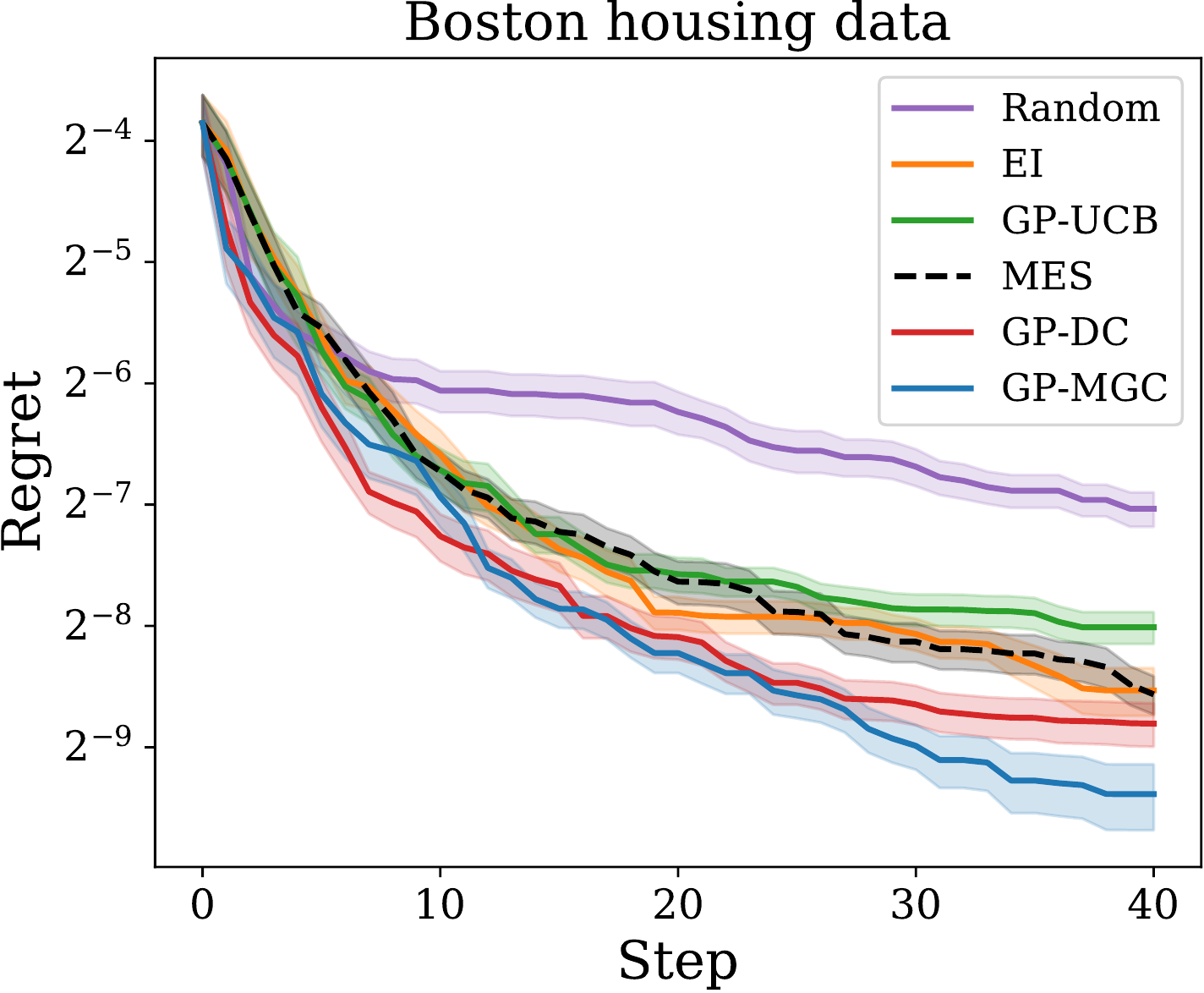}
	\quad 
	\includegraphics[width=.45\textwidth]{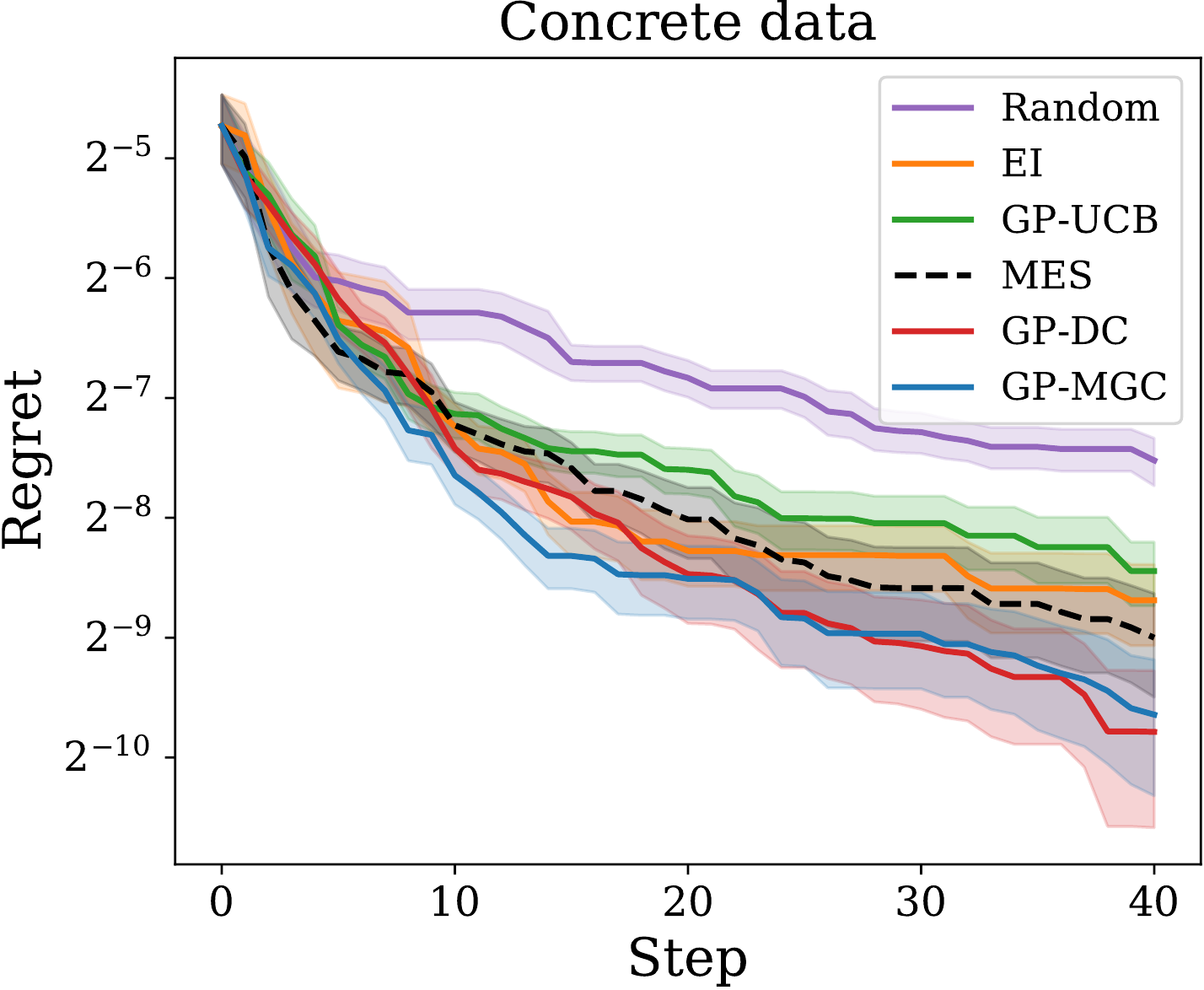}
	\\
	\centering
	\includegraphics[width=.45\textwidth]{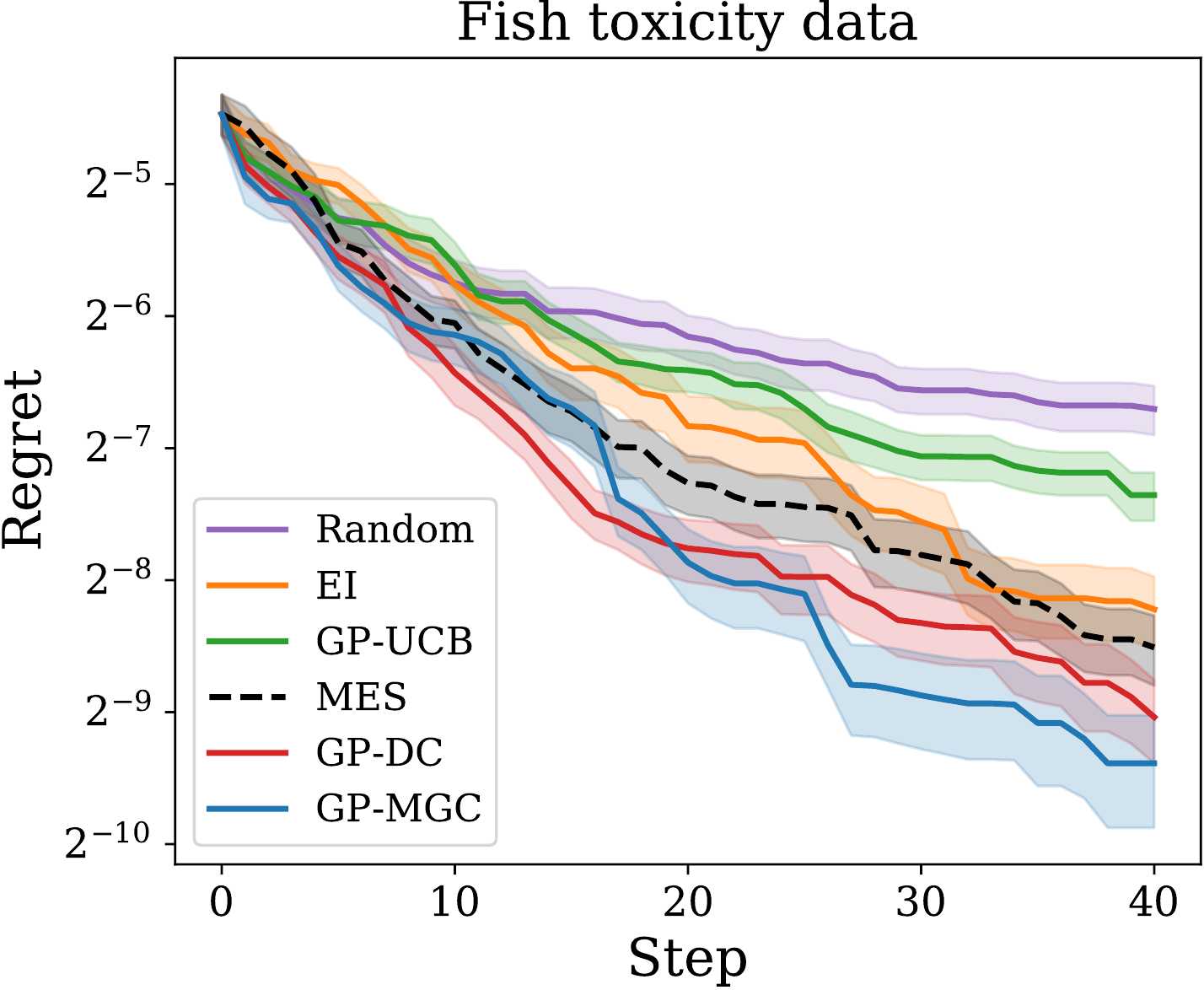}
	\caption{\label{fg:real}Benchmark results from support vector regression on three real-world datasets. Regret was computed as $0.890-\max\{R^2_t\}$, $0.901-\max\{R^2_t\}$, and $0.641-\max\{R^2_t\}$ for Boston housing data, concrete data, and fish toxicity data, respectively.}
\end{figure}

The results of numerical experiments are summarized in figure~\ref{fg:real}. GP-DC and GP-MGC outperformed the other methods, especially at steps $\gtrsim 10$. The performance of GP-DC and GP-MGC were similar. Cumulative regret is shown in table~\ref{tb:real} to provide a quantitative comparison. Overall, the results show empirically that GP-MGC was on par with GP-DC and performed better than state-of-the-art methods such as GP-UCB and MES. 

\begin{table}
	\centering 
	\begin{tabular}{|c|c|c|c|c|c|c|}
		\hline 
		& \multicolumn{2}{c|}{Boston housing} & \multicolumn{2}{c|}{Concrete} & 
		\multicolumn{2}{c|}{Fish toxicity}
		\\\hline
		Steps & 1--20 &21--40 & 1--20 &21--40 &1--20 & 21--40
		\\\hline 
		Random &0.34&0.16&0.36&0.21&0.42&0.22
		\\
		EI &0.27&0.040&0.27&0.14&0.42&0.11 
		\\
		GP-UCB &0.27&0.056&0.29&0.15&0.42&0.17 
		\\
		MES &0.27&0.042&0.26$'$&0.13&0.35&0.091 
		\\
		GP-DC &{\bf 0.17}&0.019$'$&0.27&{\bf 0.12}&{\bf 0.29}&0.066$'$
		\\
		GP-MGC &0.19$'$&{\bf 0.010}&{\bf 0.24}&0.12$'$&0.30$'$&{\bf 0.051}
		\\\hline 
	\end{tabular}
	\caption{\label{tb:real}Cumulative regret (lower is better) for the support vector machine (SVM) hyperparameter tuning task. The best score is in bold, and the second best is represented by prime $(')$.}
\end{table}

\section{Conclusions and outlook}
\label{sc:conc}
In this paper, we introduced an algorithm called GP-MGC for Bayesian optimization, which uses multiscale graph correlation (MGC) to determine where to query next on the basis of previous observations. MGC generalizes distance correlation (DC) and, as such, GP-MGC generalizes and upgrades GP-DC, which we previously proposed \cite{Kanazawa2021}. We numerically tested GP-MGC in a number of tasks involving six synthetic benchmark functions and three real-world datasets, showing that in general GP-MGC competes well with state-of-the-art methods such as GP-UCB and max-value entropy search (MES). Although GP-MGC is unlikely to outperform other algorithms in \emph{every} optimization problem \cite{noofreelunch}, we believe this study has demonstrated the practical utility of GP-MGC for solving black-box optimization problems in science and engineering such as parameter tuning of robots, molecular design, and automated machine learning. 

The findings in this study can be extended in miscellaneous directions. First, although we have only delivered an empirical study of GP-MGC, deriving an analytical guarantee for the performance of GP-MGC is desirable. Second, GP-MGC could be extended to a batch setup where multiple points are evaluated in parallel. Third, extending GP-MGC to multi-objective optimization would be highly challenging but interesting. 

\bibliography{draft_v2.bbl}
\end{document}